\documentclass[conference]{IEEEtran}
\IEEEoverridecommandlockouts
\usepackage{cite}
\usepackage{amsmath,amssymb,amsfonts}
\usepackage{algorithmic}
\usepackage{graphicx}
\usepackage{textcomp}
\usepackage{xcolor}
\usepackage{multirow}
\usepackage{array}
\usepackage{tabularx}
\usepackage{booktabs} 
\usepackage{CJKutf8}

\usepackage[framemethod=TikZ]{mdframed}
\usepackage{url}   
\usepackage{subcaption} 

\def\BibTeX{{\rm B\kern-.05em{\sc i\kern-.025em b}\kern-.08em
    T\kern-.1667em\lower.7ex\hbox{E}\kern-.125emX}}
\begin{document}
\begin{CJK}{UTF8}{gbsn}
\title{Self-Supervised Contrastive Graph Clustering Network via Structural Information Fusion}

\author{\IEEEauthorblockN{Xiaoyang Ji, Yuchen Zhou, Haofu Yang, Shiyue Xu, Jiahao Li}
\IEEEauthorblockA{\textit{Nankai University} \\
Tianjin, China \\
2110611@mail.nankai.edu.cn}
}

\maketitle
\begin{abstract}
Graph clustering, a classical task in graph learning, involves partitioning the nodes of a graph into distinct clusters. This task has applications in various real-world scenarios, such as anomaly detection, social network analysis, and community discovery. Current graph clustering methods commonly rely on module pre-training to obtain a reliable prior distribution for the model, which is then used as the optimization objective. However, these methods often overlook deeper supervised signals, leading to sub-optimal reliability of the prior distribution. To address this issue, we propose a novel deep graph clustering method called CGCN. Our approach introduces contrastive signals and deep structural information into the pre-training process. Specifically, CGCN utilizes a contrastive learning mechanism to foster information interoperability among multiple modules and allows the model to adaptively adjust the degree of information aggregation for different order structures. Our CGCN method has been experimentally validated on multiple real-world graph datasets, showcasing its ability to boost the dependability of prior clustering distributions acquired through pre-training. As a result, we observed notable enhancements in the performance of the model.
\end{abstract}

\section{Introduction}
Deep learning methods, renowned for their remarkable representation learning capabilities, have yielded promising outcomes in the realm of deep graph clustering \cite{cui2018survey, yue2022survey, liang2022reasoning}, particularly within various practical graph-based application domains like anomaly detection, social network analysis, and community discovery, etc \cite{jin2021survey, ma2022curriculum, cavallari2017learning,  MNCI_ML_SIGIR}. As a classical unsupervised task, graph clustering focuses on how to better classify the individual nodes in a graph into their corresponding clusters as much as possible without supervised signals.

The effectiveness of deep graph clustering methods heavily relies on two critical factors: the optimization objective and the feature extraction technique \cite{hamilton2017inductive, liu2022graph}. Particularly in unsupervised clustering scenarios, where the absence of label guidance poses a challenge, the design of a sophisticated objective function and an intricate feature extraction approach can significantly enhance clustering performance.

During the early stages, deep clustering methods primarily focused on leveraging attribute information within the original feature space, leading to commendable performance in numerous cases \cite{tian2014learning, grover2016node2vec}. Recent research endeavors have exhibited a strong inclination towards extracting geometric structure information and integrating it with attribute information to further enhance clustering accuracy \cite{wang2023overview, RGAE}. However, although existing methods have successfully incorporated both types of information and achieved performance gains, these methods tend to rely on simplistic fusion techniques, neglecting deeper levels of crucial data mining \cite{liu2023reinforcement}. Consequently, the reliability of prior clustering distributions obtained through pre-training various models remains insufficient.

To tackle these challenges, we propose the CGCN method. The core concept behind this framework revolves around bolstering the reliability of the pre-training's a priori distribution. We introduce a contrastive learning module \cite{li2021contrastive, pan2021multi} that facilitates cross-referencing and cross-checking among different pre-training modules, maximizing the retention of vital shared information. Furthermore, we delve into deeper graph structure information, enabling the model to dynamically and adaptively adjust the degree of aggregating different order structure information during training. 
The primary contributions of this paper can be summarized as follows:

(1) The introduction of a contrast learning mechanism fosters information interoperability among multiple modules during the pre-training process, thereby enhancing the reliability of the priori clustering distribution.

(2) The module design encourages the model to flexibly adjust the degree of information aggregation pertaining to various order structures, allowing for adaptive changes throughout the training process.

(3) We validate the effectiveness of our approach through comprehensive experiments conducted on real-world graph datasets.

\section{Related Work}

Graph neural networks (GNNs) have gained significant popularity in various graph scenarios \cite{perozzi2014deepwalk, TMac_ML_MM, yu2023g}, such as graph clustering, knowledge graph \cite{meng2023sarf}, link prediction \cite{liumeng2023self}, etc. 

Among these, attribute graph clustering is a fundamental task that poses significant challenges. It entails the segregation of nodes in an attribute graph into distinct clusters devoid of any human annotation. In early methodologies, self-encoders were employed to acquire node embeddings, which were then subjected to K-means \cite{hartigan1979algorithm} clustering.

Building upon the success of GNNs, the MGAE \cite{wang2017mgae} algorithm was introduced, which employed a graph self-encoder to encode nodes and then performed spectral clustering for node clustering. To develop a clustering-oriented approach, the DAEGC \cite{wang2019attributed} framework was proposed, incorporating an attention-based graph encoder and clustering alignment loss within deep clustering methods. SDCN \cite{bo2020structural} demonstrated the effectiveness of integrating structural and attribute information.

Contrastive learning has emerged as a research hot-spot in the field of deep graph clustering. AGE \cite{cui2020adaptive} addressed high-frequency noise in node attributes and trained the encoder using adaptive positive and negative sample comparisons. MVGRL \cite{hassani2020contrastive} generated gradual structural views and compared node embeddings with graph embeddings to enhance learning. Despite the validation of the contrast learning paradigm, several technical issues remain unresolved. GDCL \cite{zhao2021graph} aimed to correct sampling bias in deep graph clustering with contrast depth, while TGC \cite{TGC_ML} provided a generalized framework for deep node clustering in temporal graphs. To tackle the chanllege of graph scale, a scalable deep graph clustering method called S3GC \cite{devvrit2022s3gc} was proposed, leveraging comparative learning with GNNs.

Although effective, these methods usually ignore the deep structural information and the robust supervised signals form contrastive view. To address this limitation, our paper introduces a novel method called CGCN that utilizes the contrastive learning and higher-order structural information to enhance the prior clustering distribution from the pre-training process. Next, we give the details of the proposed method.

\section{Method}

\begin{figure*}[t]
\centering
\includegraphics[width=0.66\textwidth]{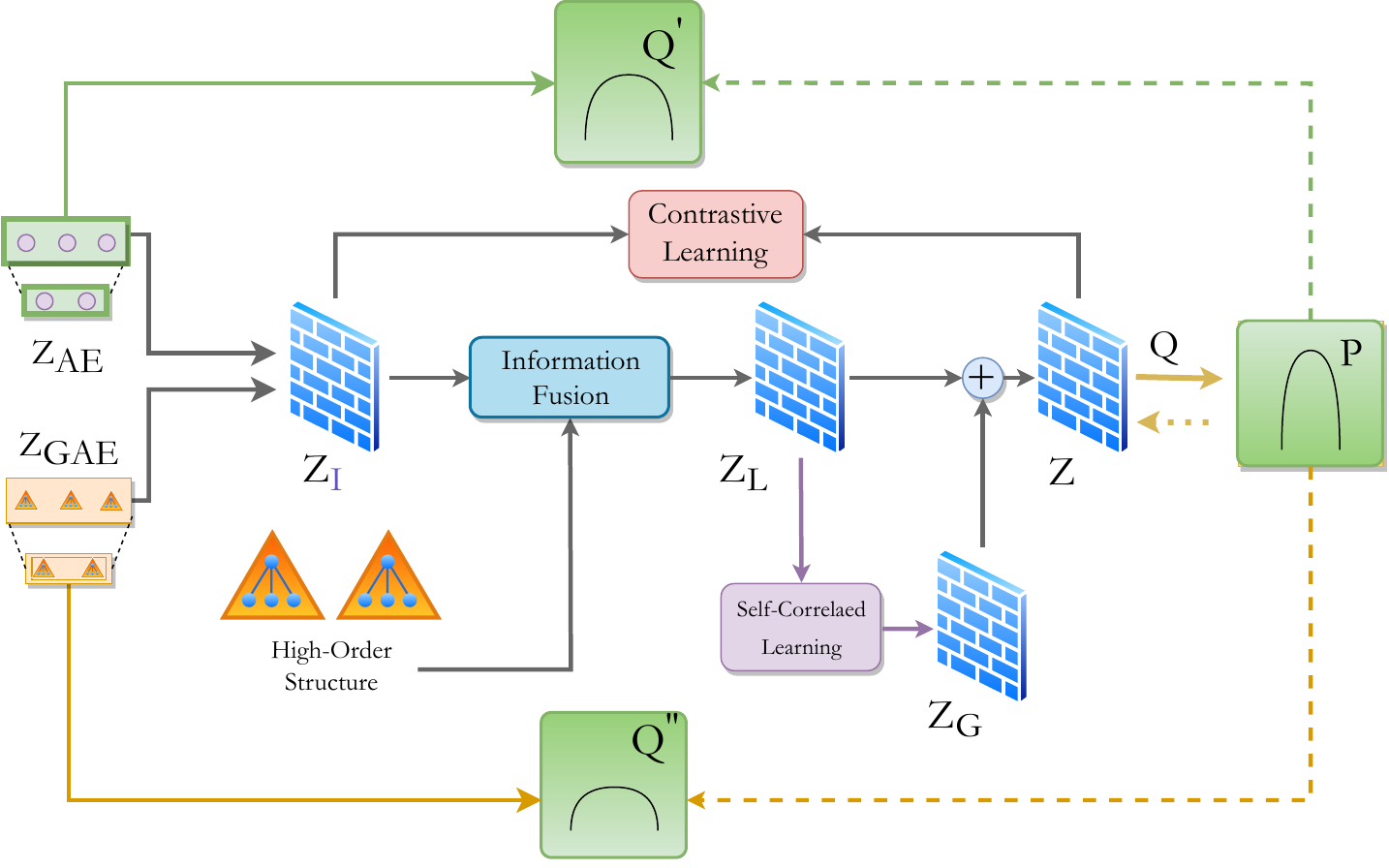}
\caption{Overall Framework}
\label{framework}
\end{figure*}

\subsection{Overall Framework}

Our proposed method, CGCN, is an improvement of the DFCN method \cite{tu2021deep}, a classical deep graph clustering method, by introducing AutoEncoder and Graph AutoEncoder to enable the computation of a priori clustering distributions during the pre-training process. We further introduce contrast learning and higher-order structural information in this session to enhance the reliability of the a priori clustering distribution. The structure of our model is shown in Fig. \ref{framework}.

\subsection{Autoencoder}

We introduce the Autoencoder (AE) as one of the pre-training modules, which is an unsupervised learning model for data reduction and feature extraction. Its goal is to learn a compact representation that can reconstruct the input data while preserving the most important features.

An autoencoder consists of two main components: an Encoder and a Decoder. The Encoder converts the input data into a low-dimensional representation (Encoding), while the Decoder maps the low-dimensional representation back to the original data space (Decoding). In this way, the autoencoder can learn a valid feature representation by minimizing the reconstruction error between the input data and the reconstructed data. Its encoder and decoder are of the form:
\begin{equation}
H = f(X), \qquad \mathbf{Z}_{AE} = g(H)
\end{equation}

Its loss function is a reconstruction of the sample features:
\begin{equation}
L_{AE} = \|\mathbf{Z}_{AE} - X\|^2_2
\end{equation}

where $X$ is the initial node features, $H$ is the output of the encoder. $\mathbf{Z}_{AE}$ is the output of the decoder, i.e. the node embeddings of AE. $f(\cdot)$ and $g(\cdot)$ are functions of the encoder and decoder, and $L_{AE}$ is the loss function used to measure the reconstruction error.

\subsection{Graph Autoencoder}

Furthermore, we hereby introduce the module of graph autoencoder (GAE) as an additional pre-training component. The primary objective of the GAE is to simultaneously reconstruct the weighted attribute matrix and the adjacency matrix. The architectural configuration of both the encoder and decoder components is detailed below.
  \begin{equation}
\mathbf{Z}^{(l)}=\sigma(\widetilde{\mathbf{A}}\mathbf{Z}^{(l-1)}\mathbf{W}^{(l)}) 
  \end{equation}
    \begin{equation}
\mathbf{Z}_{GAE} = \sigma(\widetilde{\mathbf{A}}\widehat{\mathbf{Z}}^{(h-1)}\widehat{\mathbf{W}}^{(h)})
  \end{equation}
 
$\mathbf{W}^{(l)}$ and $\mathbf{W}^{(h)}$ denote the learnable parameter matrices for the encoder and decoder layers, respectively.
The loss function for GAE is defined as follows:
    \begin{equation}
      \begin{aligned}
    L_{IGAE} = L_f + \gamma L_s
      \end{aligned}
  \end{equation}

Such loss function can be divided into two parts, i.e., feature reconstruction $L_f$ and struture reconstruction $L_s$.
\begin{equation}
\begin{aligned}
L_f = \frac{1}{2N} \|\widetilde{\mathbf{A}}\mathbf{X} - \mathbf{Z}_{GAE}\|_F^2
\label{l_f}
\end{aligned}
\end{equation}

\begin{equation}
\begin{aligned}
L_s = \frac{1}{2N} \|\widetilde{\mathbf{A}} - \widehat{\mathbf{A}}\|_F^2
\label{l_s}
\end{aligned}
\end{equation}

$\mathbf{Z}_{GAE} \in \mathbb{R}^{N \times d}$ denotes the node embeddings of GAE. Notably, Eq. (\ref{l_s}) employs inner product operations to generate the reconstructed adjacency matrix, leveraging multi-layer representations of the network. By minimizing both Eq. (\ref{l_f}) and Eq. (\ref{l_s}), the GAE module aims to effectively reduce the reconstruction loss associated with the weighted attribute matrix and the adjacency matrix.

\subsection{Structural and Attribute Information Fusion}

After pre-training the AE and GAE to obtain their respective node embeddings, we will use these embeddings to generate the final node embeddings as well as the priori clustering distributions. In this subsection, we first present the embedding generation, i.e., structural and attribute information fusion. The initial fusion embeddings $\mathbf{Z}_I$ can be calculated as follows.
\begin{equation}
\begin{aligned}
\mathbf{Z}_I=\delta\mathbf{Z}_{AE}+(1-\delta)\mathbf{Z}_{IGAE}
  \end{aligned}
  \end{equation}

To process the combined information $\mathbf{Z}_{I}$, we employ an operation similar to graph convolution. This operation allows us to enhance the initial fused embeddings by incorporating the local structure present in the data. The combined information $\mathbf{Z}_I$ is represented as $\mathbf{Z}_I\in\mathbb{R}^{N\times d^{\prime}}$, where $d^{\prime}$ represents the dimensionality of the enhanced embeddings.
  \begin{equation}
  \begin{aligned}
  \mathbf{Z}_L=\tilde{\mathbf{A}}\mathbf{Z}_I
  \end{aligned}
  \end{equation}

Here, $\mathbf{Z}^1\in\mathbb{R}^{N\times d^{\prime}}$ denotes the locally enhanced representation from the first-order neighborhood. In this way, we can combine different order neighborhood information by the adjacency matrix. Numerous experiments have proved that graph neural networks with more than three layers can bring about over-smoothing phenomenon, so we only select the first two layers of neighborhood information to be combined here.
  \begin{equation}
  \begin{aligned}
\mathbf{Z}_L=\lambda_1 \mathbf{Z}^1 + \lambda_2 \mathbf{Z}^2
  \end{aligned}
  \end{equation}

$\lambda_1$ and $\lambda_2$ are learnable parameters that control the fusion degree of different order neighborhood information.

Consequently, we propose the incorporation of a self-correlation mechanism to leverage non-local relationships within the initial information fusion space among samples. To this end, we first calculate the normalized self-correlation matrix denoted as $\mathbf{S}\in\mathbb{R}^{N\times N}$, as described by Eq. (\ref{S}).
\begin{equation}
\begin{aligned}
\mathbf{S}_{ij}=\frac{e^{(\mathbf{Z}_{L}\mathbf{Z}_{L}^{\mathbf{T}})_{ij}}}{\sum_{k=1}^{N}e^{(\mathbf{Z}_{L}\mathbf{Z}_{L}^{\mathbf{T}})_{ik}}}
\label{S}
  \end{aligned}
  \end{equation}

Following this, considering S as the coefficient, we take into account the global inter-sample correlations and perform a recombination.
 \begin{equation}
\mathbf{Z}_G=\tilde{\mathbf{S}}\mathbf{Z}_L
  \end{equation}

Lastly, we employ skip connections to facilitate the smooth transmission of information within the fusion mechanism.
 \begin{equation}
  \begin{aligned}
\mathbf{Z}_{final}=\lambda_b \mathbf{Z}_G+\mathbf{Z}_L
  \end{aligned}
  \end{equation}

Here, the scaling parameter $\lambda_b$ is introduced to adjust the contribution of the cross-modal dynamic fusion mechanism. This mechanism effectively considers both local and global levels of sample correlations. By intricately integrating and refining the information obtained from the Autoencoder (AE) and Graph Autoencoder (GAE), this method facilitates a more accurate learning of consensus latent representations.

\subsection{Priori Clustering Distribution}

To provide more reliable guidance during the training of the clustering network, the model initially employs a robust clustering embedding $\widetilde{\mathbf{Z}}\in\mathbb{R}^{N\times d'}$. This embedding integrates information from both the Autoencoder (AE) and Improved Graph Autoencoder (IGAE) to facilitate target distribution generation. The generation process can be formulated as follows.
\begin{equation}
\begin{aligned}
q_{ij}&=\frac{(1+\|\tilde{z}_i-u_j\|^2/v)^{-\frac{v+1}2}}{\sum_{j^{\prime}}(1+\|\tilde{z}_i-u_{j^{\prime}}\|^2/v)^{-\frac{v+1}2}}
\label{q}
  \end{aligned}
  \end{equation}
\begin{equation}
\begin{aligned}p_{ij}&=\frac{q_{ij}^2/\sum_iq_{ij}}{\sum_{j^{\prime}}(q_{ij^{\prime}}^2/\sum_iq_{ij^{\prime}})}
\label{p}
  \end{aligned}
  \end{equation}




In the initial stage, the similarity between the $i$-th sample ($\tilde{z}_{i}$) and the $j$-th pre-calculated clustering center ($u_j$) in the fused embedding space is computed using the Student's t-distribution as a kernel. This process is represented by Equation (\ref{q}), where $v$ denotes the degree of freedom for the Student's t-distribution and $q_{ij}$ represents the probability of assigning the $i$-th node to the $j$-th center, indicating a soft assignment.

The soft assignment matrix $\mathbf{Q}\in\mathbb{R}^{N\times K}$ reflects the distribution of all samples. In the subsequent step, Equation (\ref{p}) is introduced to encourage all samples to move closer to the cluster centers, thereby enhancing the confidence of cluster assignment. Specifically, $0\leq p_{ij}\leq1$ denotes an element of the generated target distribution $\mathbf{P}\in\mathbb{R}^{N\times K}$, indicating the probability of the $i$-th sample belonging to the $j$-th cluster center.

Using the iteratively generated target distribution, the soft assignment distribution of the AE and GAE modules is calculated separately by applying Equation (\ref{q}) to the latent embeddings of each module. The soft assignment distributions of the AE and GAE are denoted as $Q^{\prime}$ and $Q^{\prime\prime}$, respectively.

To achieve integration of network training and enhanced representation capacity for each component, we introduce a triplet clustering loss that utilizes adapted KL-divergence. The proposed loss function can be defined as follows.
  \begin{equation} \begin{aligned}
  L_{KL}=\sum_{i}\sum_{j}p_{ij}log\frac{p_{ij}}{(q_{ij}+q_{ij}^{'}+q_{ij}^{''})/3}
  \end{aligned}
  \end{equation}

In this formulation, we aim to align the summation of the soft assignment distributions of the two modules, and the fused representations with the robust target distribution. As the target distribution is generated without human guidance, we refer to the loss function as the triplet clustering loss and the corresponding training mechanism as the triplet self-supervised strategy.

\subsection{Loss Function}
The overall learning objective consists of two main parts, i.e., the reconstruction loss of AE and GAE, and the clustering loss which is correlated with the target distribution:
\begin{equation}
\begin{aligned}
L = L_{AE} + L_{GAE} + L_C + \lambda L_{KL}
  \end{aligned}
  \end{equation}

$\lambda$ is a pre-defined hyper-parameter which balances the importance of reconstruction and clustering.

$L_C$ is a loss function that introduces the contrastive learning, which aims to enhance the constrative signals for the model training. In particular, $L_C$ can be divided into two parts: pre-training embedding alignment and training embedding alignment.
\begin{equation}
L_C = \alpha L_{pre} + \beta L_{train}
  \end{equation}
\begin{equation}
L_{pre} = \|\mathbf{Z}_{GAE} - \mathbf{Z}_{AE}\|^2_2
  \end{equation}
\begin{equation}
L_{train} = \|\mathbf{Z}_{final} - \mathbf{Z}_{AE}\|^2_2
  \end{equation}

By incorporating these constraints into the contrast mechanism, we aim to encourage the model to acquire more trustworthy contrast signals throughout the pre-training and training stages. This ensures that the optimization process progresses in a more accurate direction, leading to improved model performance.

\section{Experiments}

\begin{table}[t]
\centering
\caption{Dataset Description.}
\label{dataset}
\begin{tabular}{c|cccc}
\toprule[2pt]
Datasets              & Nodes & Edges & Clusters& Feature \\
\midrule[1pt]
DBLP  & 4,058    & 3,528 &4    & 334\\
CITE  & 3,327    & 4,732 &6    & 3,703\\
ACM  & 3,025    & 13,128 &3    & 1,870\\
\bottomrule[2pt]
\end{tabular}
\end{table}

\begin{table*}[t]
\centering
\caption{Node Clustering Performance. We bold the best results and add underline to the second best results.}
\label{nc}
\resizebox{1\textwidth}{22mm}{
\begin{tabular}{c|c|ccccccccccc|cc}
\toprule[2pt]
Datasets              & Mteric & Kmeans & AE   & DEC  & IDEC & GAE  & VGAE & ARGA & DAEGC & SDCNQ & SDCN & DFCN & CGCN & Improv.  \\
\midrule[1pt]
\multirow{4}{*}{DBLP} 
&ACC&38.7$\pm$0.7&51.4$\pm$0.4&58.2$\pm$0.6&60.3$\pm$0.6&61.2$\pm$1.2&58.6$\pm$0.1&61.6$\pm$1.0&62.1$\pm$0.5&65.7$\pm$1.3&68.1$\pm$1.8&\underline{76.0$\pm$0.8}&\textbf{77.3$\pm$0.2}&+1.71\%\\
&NMI&11.5$\pm$0.4&25.4$\pm$0.2&29.5$\pm$0.3&31.2$\pm$0.5&30.8$\pm$0.9&26.9$\pm$0.1&26.8$\pm$1.0&32.5$\pm$0.5&35.1$\pm$1.1&39.5$\pm$1.3&\underline{43.7$\pm$1.0}&\textbf{45.2$\pm$0.4}&+3.43\%\\
&ARI&7.0$\pm$0.4&12.2$\pm$0.4&23.9$\pm$0.4&25.4$\pm$0.6&22.0$\pm$1.4&17.9$\pm$0.1&22.7$\pm$0.3&21.0$\pm$0.5&34.0$\pm$1.8&39.2$\pm$2.0&\underline{47.0$\pm$1.5}&\textbf{49.1$\pm$0.5}&+4.46\%\\
&F1&31.9$\pm$0.3&52.5$\pm$0.4&59.4$\pm$0.5&61.3$\pm$0.6&61.4$\pm$2.2&58.7$\pm$0.1&61.8$\pm$0.9&61.8$\pm$0.7&65.8$\pm$1.2&67.7$\pm$1.5&\underline{75.7$\pm$0.8}&\textbf{76.9$\pm$0.6}&+1.58\%\\
\midrule[1pt]
\multirow{4}{*}{CITE} 
&ACC&39.3$\pm$3.2&57.1$\pm$0.1&55.9$\pm$0.2&60.5$\pm$1.4&61.4$\pm$0.8&61.0$\pm$0.4&56.9$\pm$0.7&64.5$\pm$1.4&61.7$\pm$1.1&66.0$\pm$0.3&\underline{69.5$\pm$0.2}&\textbf{70.3$\pm$0.1}&+1.16\%\\
&NMI&16.9$\pm$3.2&27.6$\pm$0.1&28.3$\pm$0.3&27.2$\pm$2.4&34.6$\pm$0.7&32.7$\pm$0.3&34.5$\pm$0.8&36.4$\pm$0.9&34.4$\pm$1.2&38.7$\pm$0.3&\underline{43.9$\pm$0.2}&\textbf{44.7$\pm$0.2}&+1.83\%\\
&ARI&13.4$\pm$3.0&29.3$\pm$0.1&28.1$\pm$0.4&25.7$\pm$2.7&33.6$\pm$1.2&33.1$\pm$0.5&33.4$\pm$1.5&37.8$\pm$1.2&35.5$\pm$1.5&40.2$\pm$0.4&\underline{45.5$\pm$0.3}&\textbf{46.5$\pm$0.1}&+2.20\%\\
&F1&36.1$\pm$3.5&53.8$\pm$0.1&52.6$\pm$0.2&61.6$\pm$1.4&57.4$\pm$0.8&57.7$\pm$0.5&54.8$\pm$0.8&62.2$\pm$1.3&57.8$\pm$1.0&63.6$\pm$0.2&\underline{64.3$\pm$0.2}&\textbf{65.1$\pm$0.1}&+1.25\%\\
\midrule[1pt]
\multirow{4}{*}{ACM}
&ACC&67.3$\pm$0.7&81.8$\pm$0.1&84.3$\pm$0.8&85.1$\pm$0.5&84.5$\pm$1.4&84.1$\pm$0.2&86.1$\pm$1.2&86.9$\pm$2.8&87.0$\pm$0.1&90.5$\pm$0.2&\underline{90.9$\pm$0.2}&\textbf{91.5$\pm$0.1}&+0.67\%\\
&NMI&32.4$\pm$0.5&49.3$\pm$0.2&54.5$\pm$1.5&56.6$\pm$1.2&55.4$\pm$1.9&53.2$\pm$0.5&55.7$\pm$1.4&56.2$\pm$4.2&58.9$\pm$0.2&68.3$\pm$0.3&\underline{69.4$\pm$0.4}&\textbf{70.3$\pm$0.2}&+1.30\%\\
&ARI&30.6$\pm$0.7&54.6$\pm$0.2&60.6$\pm$1.9&62.2$\pm$1.5&59.5$\pm$3.1&57.7$\pm$0.7&62.9$\pm$2.1&59.4$\pm$3.9&65.3$\pm$0.2&73.9$\pm$0.4&\underline{74.9$\pm$0.4}&\textbf{75.6$\pm$0.2}&+0.94\%\\
&F1&67.6$\pm$0.7&82.0$\pm$0.1&84.5$\pm$0.7&85.1$\pm$0.5&84.7$\pm$1.3&84.2$\pm$0.2&86.1$\pm$1.2&87.1$\pm$2.8&86.8$\pm$0.1&90.4$\pm$0.2&\underline{90.8$\pm$0.2}&\textbf{91.1$\pm$0.2}&+0.34\% \\
\bottomrule[2pt]
\end{tabular}}
\end{table*}

\begin{figure*}[htp]
    \centering
\begin{minipage}[t]{0.3\textwidth}
\includegraphics[width=1\textwidth]{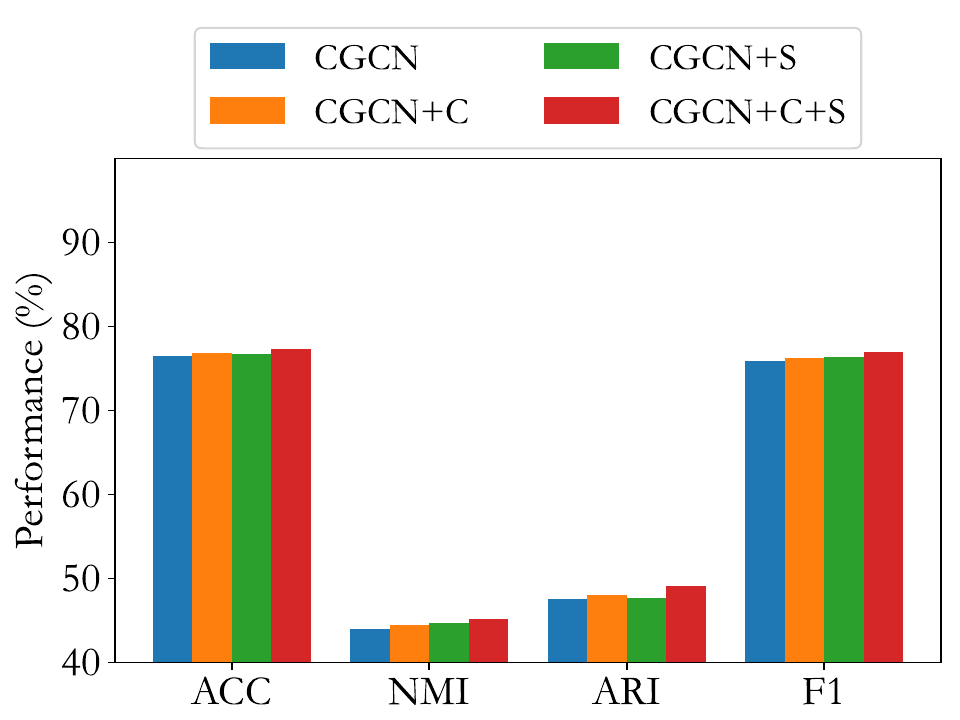}
\centerline{(a) DBLP}

    \end{minipage}%
            \begin{minipage}[t]{0.3\textwidth}
\includegraphics[width=1\textwidth]{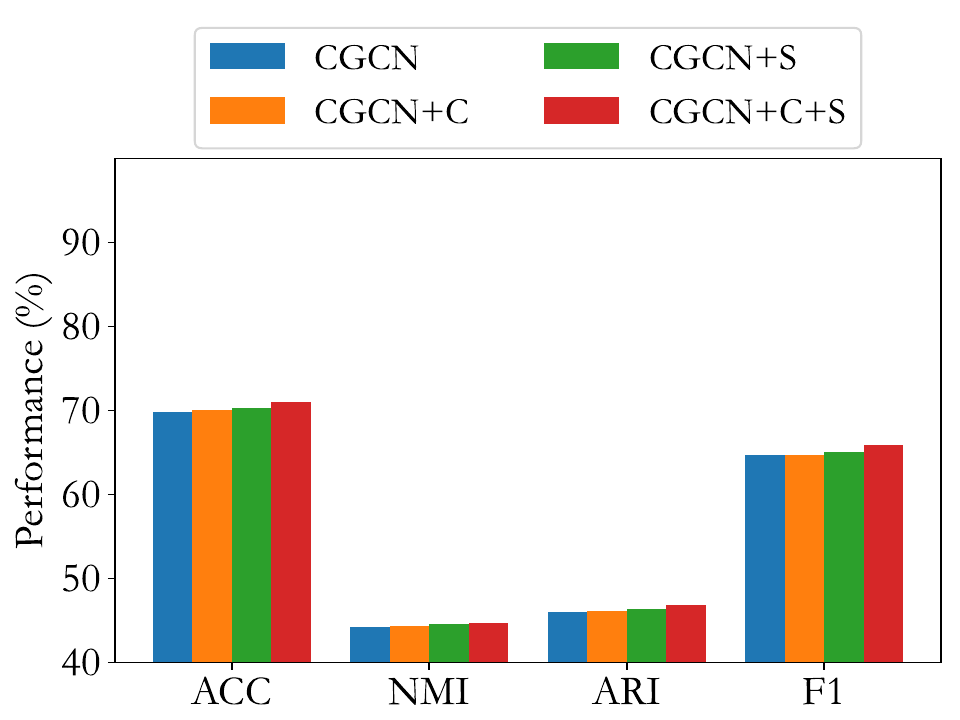}
        \centerline{(b) CITE}

    \end{minipage}%
            \begin{minipage}[t]{0.3\textwidth}
\includegraphics[width=1\textwidth]{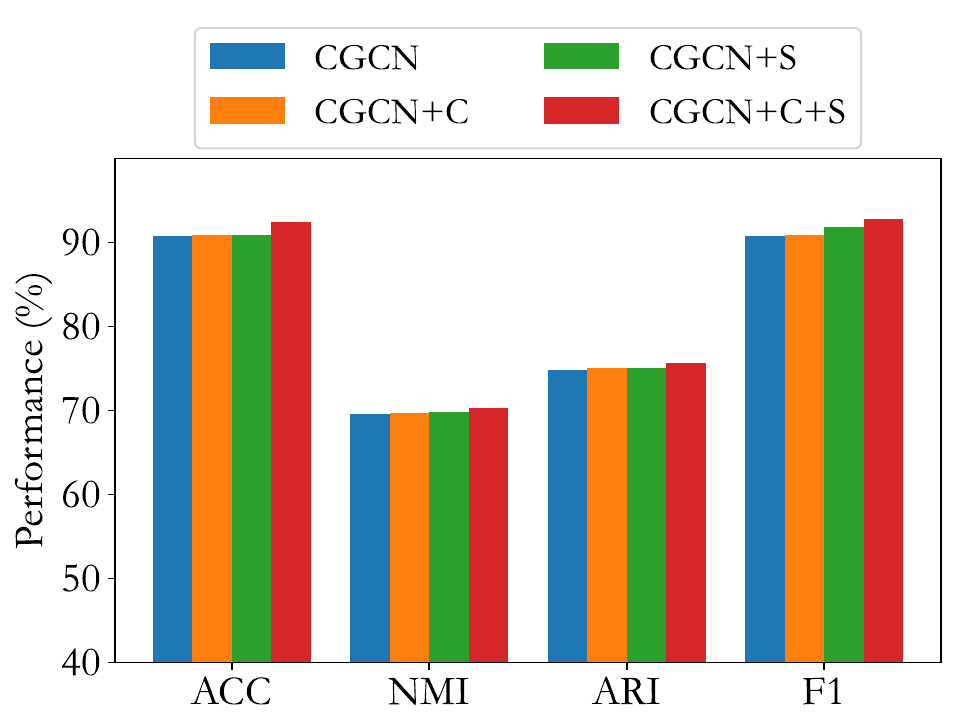}
\centerline{(c) ACM}

    \end{minipage}%
\caption{Ablation Study.}
\end{figure*}

\begin{figure}[t]
    \centering
\begin{minipage}[t]{0.16\textwidth}
\includegraphics[width=1\textwidth]{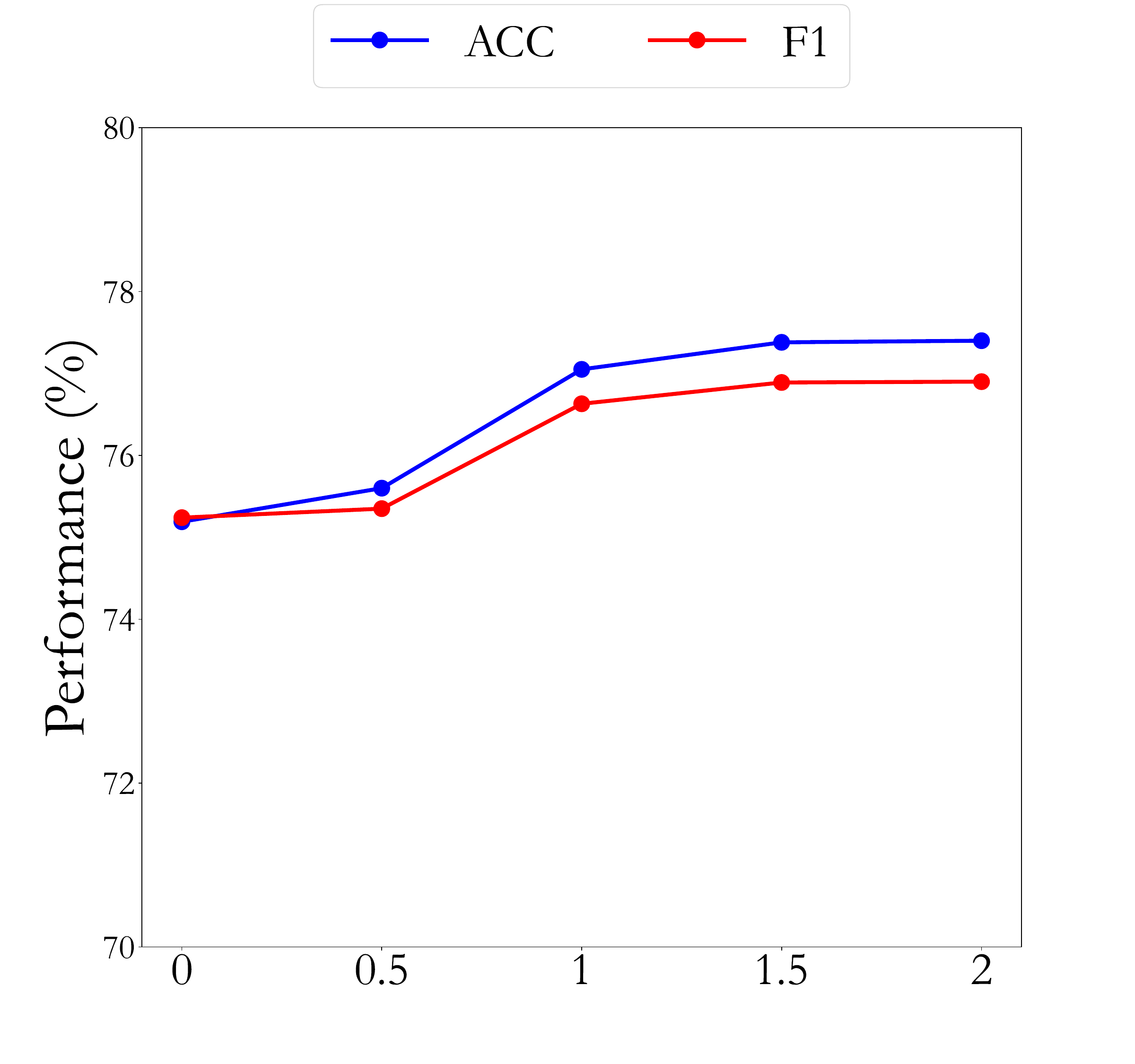}
        \centerline{(a) DBLP with $\alpha$}

    \end{minipage}%
            \begin{minipage}[t]{0.16\textwidth}
\includegraphics[width=1\textwidth]{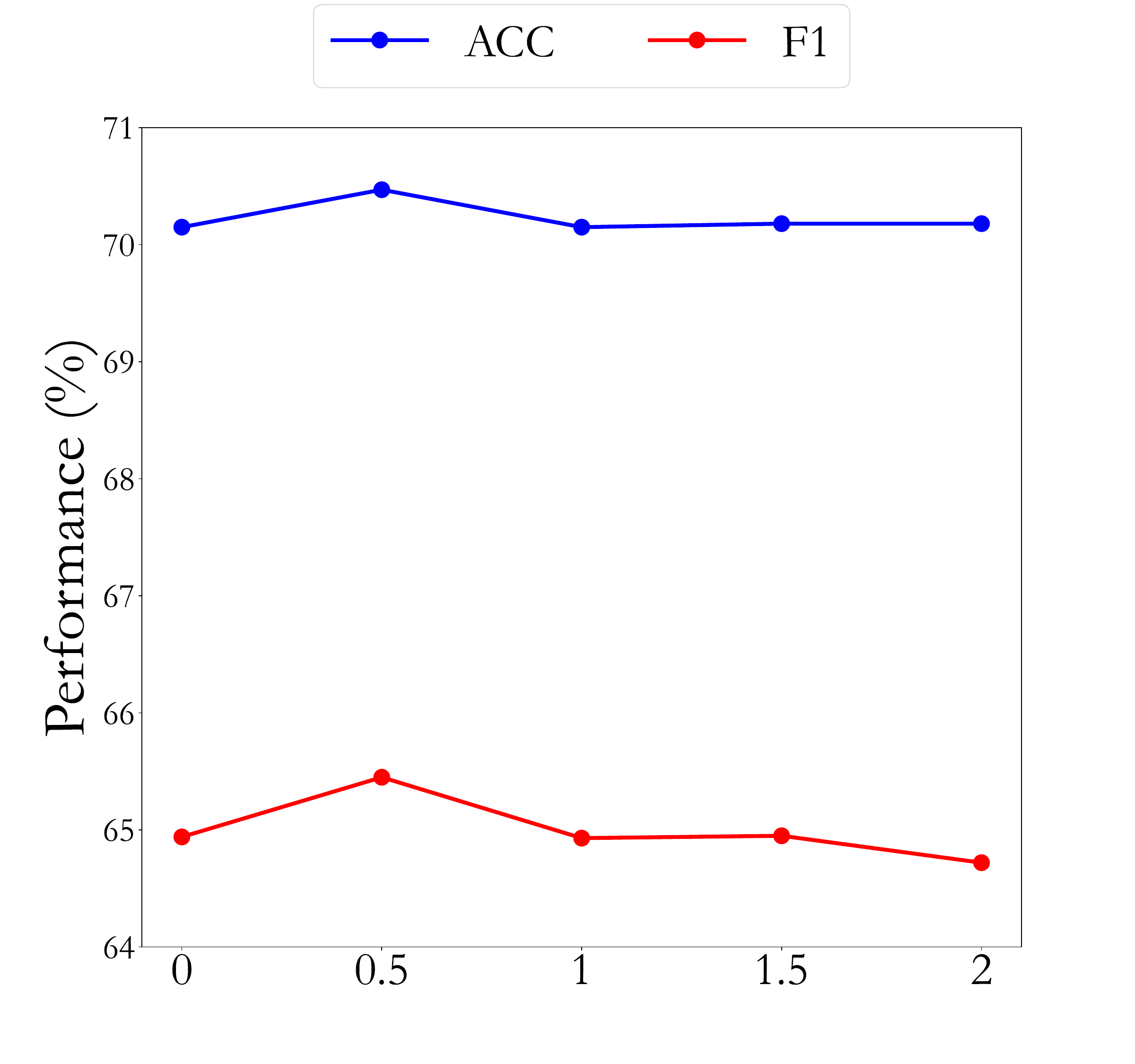}
        \centerline{(c) CITE with $\alpha$}

    \end{minipage}%
            \begin{minipage}[t]{0.16\textwidth}
        \includegraphics[width=1\textwidth]{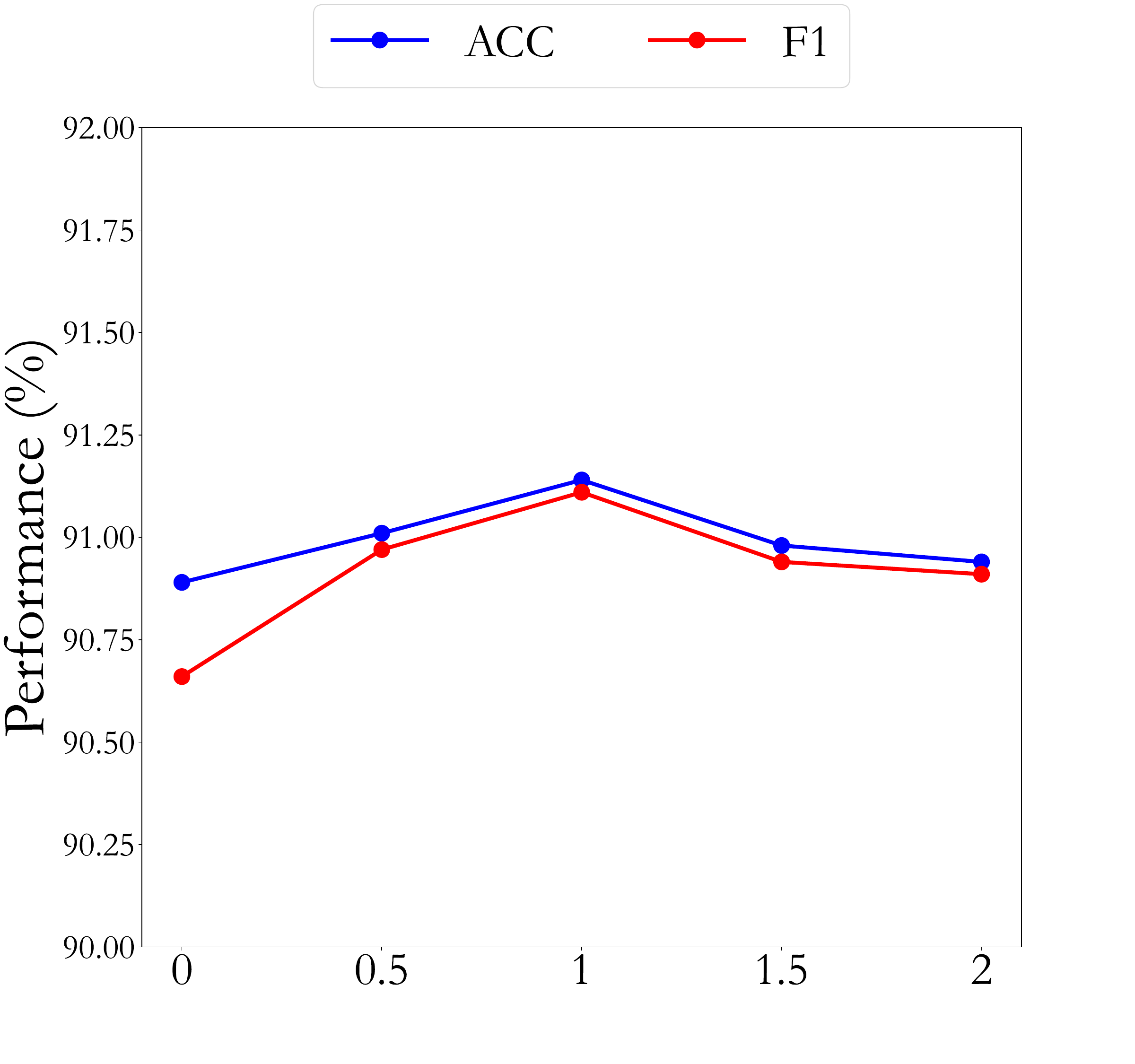}
        \centerline{(e) ACM with $\alpha$}

    \end{minipage}%

    \begin{minipage}[t]{0.16\textwidth}
\includegraphics[width=1\textwidth]{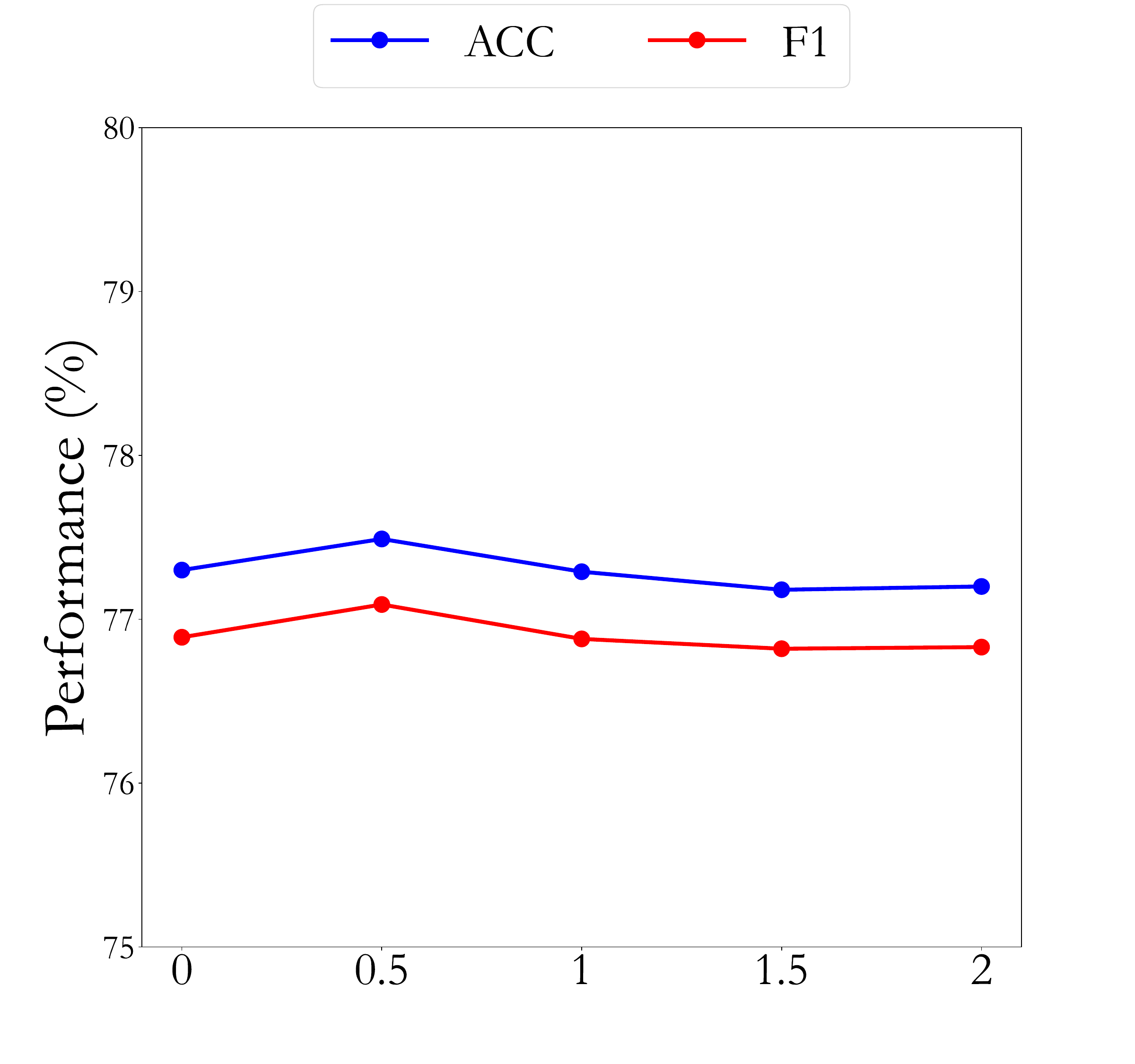}
        \centerline{(b) DBLP with $\beta$}

    \end{minipage}%
            \begin{minipage}[t]{0.16\textwidth}
    \includegraphics[width=1\textwidth]{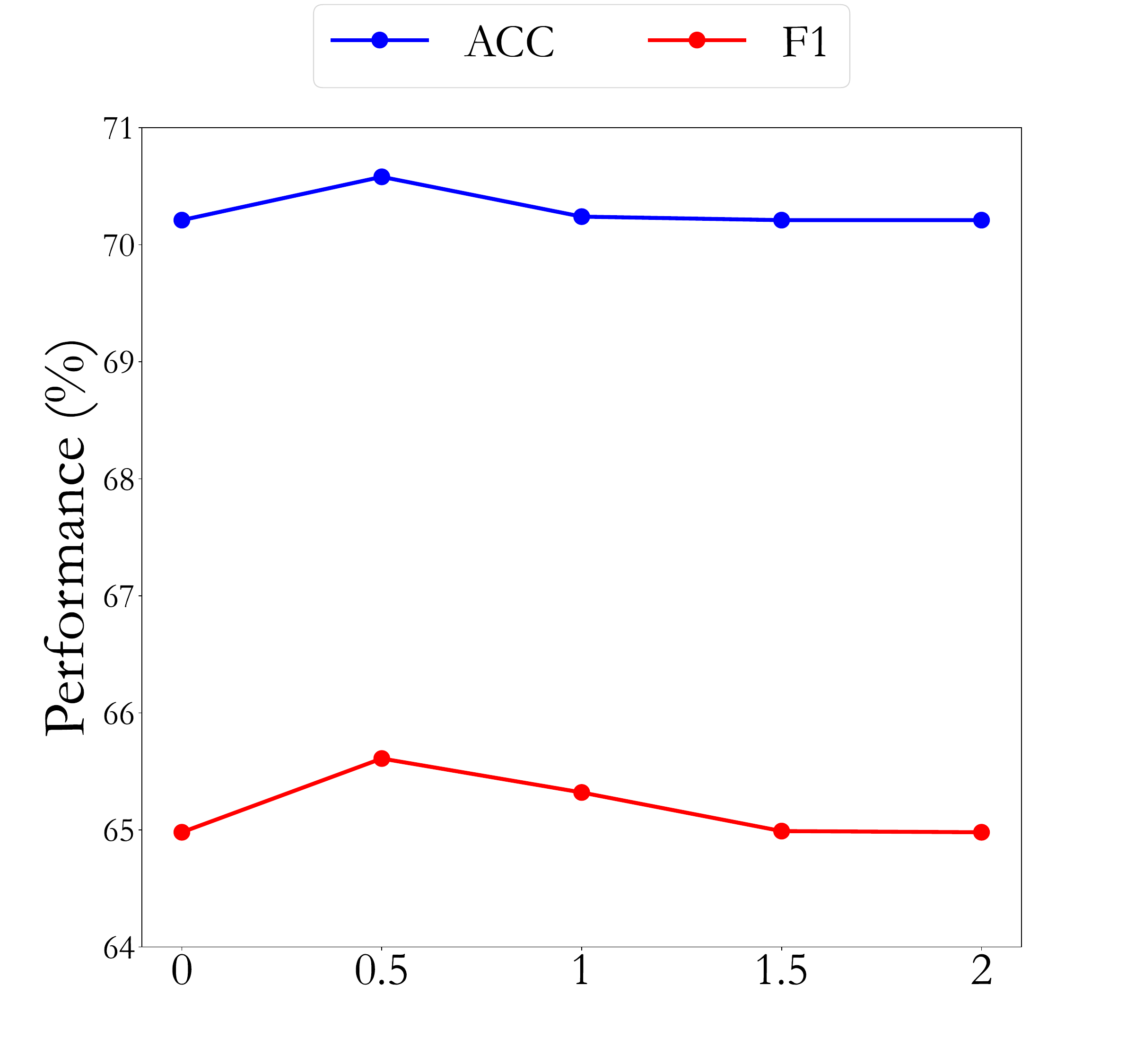}
        \centerline{(d) CITE with $\beta$}

    \end{minipage}%
            \begin{minipage}[t]{0.16\textwidth}
    \includegraphics[width=1\textwidth]{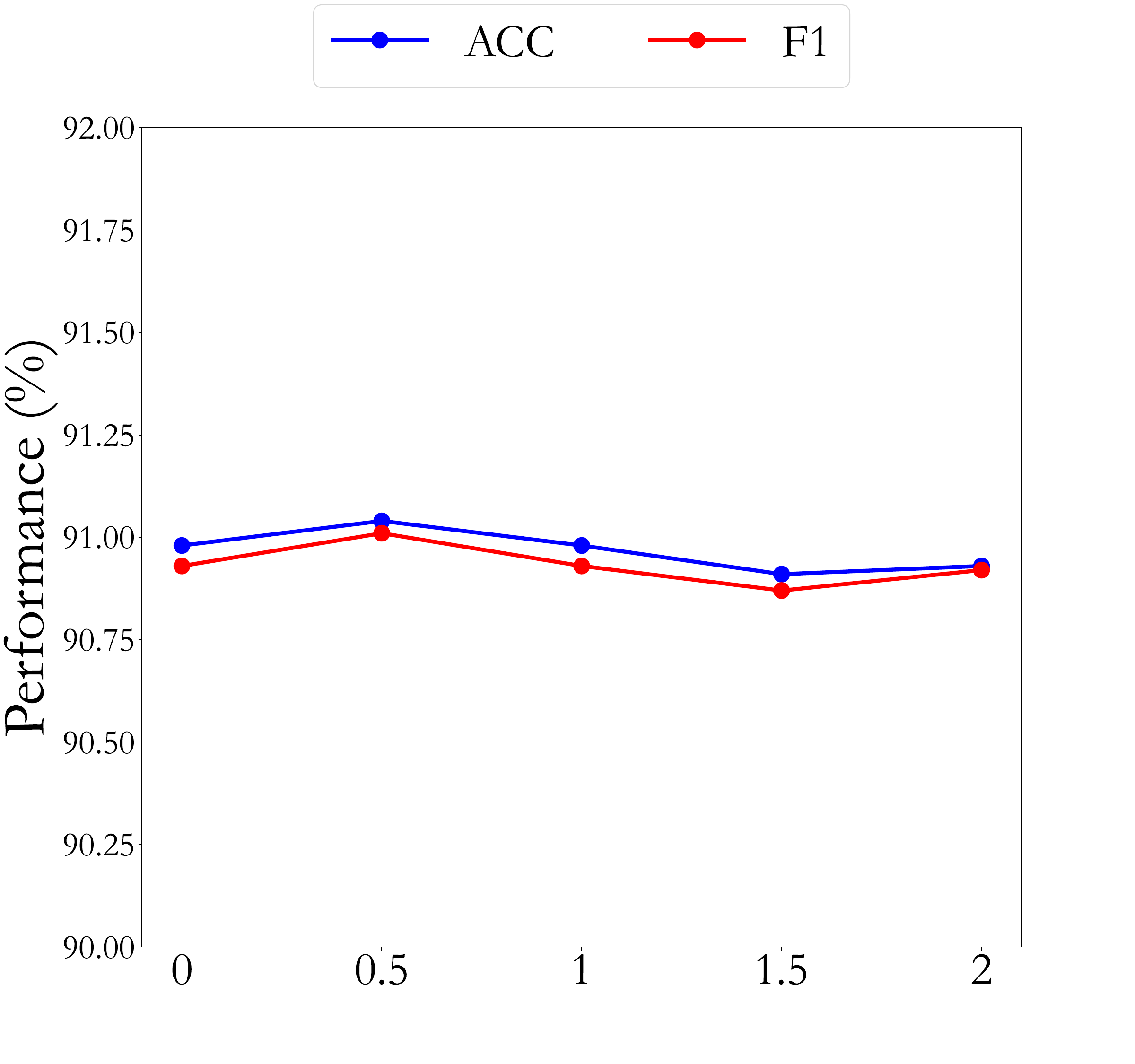}
        \centerline{(f) ACM with $\beta$}

    \end{minipage}%

    \caption{Parameter sensitivity study.}
    
\end{figure}

\subsection{Datasets and Baselines}

We conduct experiments on several real-world datasets, i.e., DBLP, CITE, and ACM. As shown in Table \ref{dataset}, these datasets have different edge numbers and feature dimensions.

To showcase the effectiveness of our proposed method, we conduct a comparative analysis with several state-of-the-art (SOTA) methods. These methods include AE \cite{hinton2006reducing}, DEC \cite{xie2016unsupervised}, and IDEC \cite{guo2017improved}, which are autoencoder-based clustering approaches that learn representations for clustering by training an autoencoder. Additionally, GAE and VGAE \cite{kipf2016variational}, ARGA \cite{pan2019learning}, and DAEGC \cite{wang2019attributed} are representative graph convolutional network-based methods. These approaches embed the clustering representation with structural information using Graph Convolutional Networks (GCN). Furthermore, SDCN \cite{bo2020structural} and DFCN \cite{tu2021deep} are hybrid methods that leverage both AE and GCN modules for clustering.

\subsection{Experimental Settings}
For comparison baseline methods, we use their default parameters. For our model, in the part of loss function, we use two parameters $\alpha$ and $\beta$ to control the pre-training loss and training loss which introduce comparative learning, respectively. For different data sets, the values of these two hyper-parameters are different, which will be introduced in detail in the parameter sensitivity experiment.

\subsection{Node Clustering Performance}
The clustering performance of our proposed CGCN method and the baseline methods on three benchmark datasets is summarized in Table \ref{nc}. Based on the results, we make the following observations:

\begin{itemize}
\item[*] CGCN consistently outperforms the compared methods in most cases. K-means performs clustering on raw data, while AE, DEC, and IDEC only utilize node attribute representations for clustering, neglecting the structural information. SDCN and DFCN rely on simplistic fusion techniques, which do not fully capture the deeper levels of important data patterns. As a result, the reliability of the prior clustering distributions obtained through pre-training various models remains insufficient.

\item[*] 
GCN-based techniques like GAE, VGAE, ARGA and DAEGC lack the same level of effectiveness demonstrated by CGCN. These approaches inadequately exploit the information inherent in the data and may be subject to constraints related to oversmoothing. In contrast, CGCN integrates attribute-based representations acquired through AE within the comprehensive clustering framework. This integration enables concurrent exploration of graph structure and node attributes, enabled by a fusion module that facilitates consensus representation learning. Consequently, CGCN delivers a substantial enhancement in clustering performance in comparison to existing GCN-based methods.

\item[*] CGCN achieves better clustering results than the strongest baseline method, DFCN, in the majority of cases, particularly on the DBLP, ACM, and CITE datasets. For example, on the DBLP dataset, CGCN achieves a 1.71\% improvement in accuracy (ACC), a 3.43\% improvement in normalized mutual information (NMI), a 4.46\% improvement in adjusted Rand index (ARI), and a 1.58\% improvement in F1 score compared to DFCN. This improvement is attributed to CGCN's contrast learning mechanism, which promotes information interoperability among multiple modules during pre-training, enhancing the reliability of the prior clustering distribution. Furthermore, CGCN allows for flexible adjustment of the degree of information aggregation related to different order structures, enabling adaptive changes throughout the training process.
\end{itemize}

\subsection{Ablation Study}

To further validate the effectiveness of our proposed method, we conducted ablation studies. In these studies, we compared the performance of DFCN with two modified versions: DFCN+C and DFCN+S. DFCN+C represents the introduction of only the contrastive learning mechanism, while DFCN+S denotes adjusting the degree of information aggregation related to various order structure information. 

Across all three datasets, we consistently observed that both DFCN+C and DFCN+S outperformed the baseline DFCN. The improvement in clustering performance was more significant when the contrastive learning mechanism was introduced, compared to solely adjusting the degree of information aggregation related to various order structure information. Moreover, when both mechanisms were employed simultaneously in DFCN+C+S, even better results were achieved. These findings collectively highlight that our proposed method can leverage more comprehensive information to enhance the generalization capabilities of deep clustering networks.

\subsection{Parameter Sensitivity}

The introduction of two hyperparameters, $\alpha$ and $\beta$, in CGCN allows us to control the proportion of dissimilarity between $Z_{AE}$ and $\tilde{\mathbf{A}}$ and between $Z_{AE}$ and $Z_{IGAE}$ during training. To investigate the impact of these parameters on all datasets, we conducted experiments by fixing the value of one parameter and varying the value of the other parameter from 0 to 2 in increments of 0.5. The results, as shown in the figures, provide the following insights:

\begin{itemize}

\item[*] The hyper-parameters $\alpha$ and $\beta$ have a significant effect on improving the clustering performance, and the adjustment of their ratio yields different clustering outcomes.

\item[*] The method exhibits stable performance changes within the range of values for $\alpha$ and $\beta$.

\item[*] The optimal ratios of the two hyper-parameters vary for different datasets. For the ACM dataset, the best model performance is achieved when $\alpha$:$\beta$ = 0.5:1. For the CITE dataset, the optimal parameters are $\alpha$:$\beta$ = 0.5:0.5. Lastly, for the DBLP dataset, the optimal parameters are $\alpha$:$\beta$ = 2:0.5. These findings suggest that the optimal ratio of the hyper-parameters depends on the specific characteristics and complexities of each dataset.
\end{itemize}

\section{Conclusion}
In order to overcome the limitations of neglecting deep contrast and structural information in depth graph clustering, we present a novel method called Contrastive Graph Convolutional Network (CGCN). Our approach integrates contrast signals and deep structural information into the pre-training phase. Experimental evaluations conducted on various real graph datasets demonstrate that our proposed CGCN method enhances the reliability of the prior clustering distributions obtained from pre-training, leading to improved model performance. Moving forward, our future research endeavors will focus on developing scalable graph clustering frameworks to accommodate larger-scale graph datasets.

\bibliographystyle{IEEEtran}
\bibliography{bib}
\end{CJK}
\end{document}